\documentclass[11pt]{article}

\usepackage[preprint]{acl}
\usepackage{booktabs}
\usepackage{times}
\usepackage{latexsym}
\usepackage{longtable}
\usepackage{dblfloatfix}
\usepackage{placeins}    % for \FloatBarrier
\usepackage{makecell}    % 
\usepackage{array}     % for \newcolumntype
\usepackage{tabularx}
\newcolumntype{Y}{>{\raggedright\arraybackslash}X}
\newcolumntype{P}[1]{>{\raggedright\arraybackslash}p{#1}} % fixed-width, wrapped

\usepackage{tabularx}
\newcolumntype{Y}{>{\raggedright\arraybackslash}X}

\usepackage[T1]{fontenc}
\usepackage[utf8]{inputenc}

\usepackage{microtype}

\usepackage{inconsolata}

\usepackage{graphicx}

\title{Retrieval--Reasoning Processes for Multi-hop Question Answering: A Four-Axis Design Framework and Empirical Trends}
\author{Yuelyu Ji \\
  School of Computing and Information\\
  University of Pittsburgh \\
  \texttt{yueluji@gmail.com} \\\And
  Zhuochun Li \\
  School of Computing and Information  \\
  University of Pittsburgh \\ \And
  Daqing He \\
  School of Computing and Information  \\
  University of Pittsburgh \\
  \texttt{dah44@pitt.edu} \\ \And}

\author{
 \textbf{Yuelyu Ji\textsuperscript{1}},
  \textbf{Zhuochun Li\textsuperscript{1}},
 \textbf{Rui Meng\textsuperscript{2}},
 \textbf{Daqing He\textsuperscript{1}}
\\
\\
 \textsuperscript{1}School of Computing and Information, University of Pittsburgh,
 \textsuperscript{2}Google Cloud AI Research
\\
\\
}
\begin{document}
\maketitle
\begin{abstract}
Multi-hop question answering (QA) requires systems to iteratively retrieve evidence and reason across multiple hops. While recent RAG and agentic methods report strong results, the underlying retrieval--reasoning \emph{process} is often left implicit, making procedural choices hard to compare across model families. This survey takes the execution procedure as the unit of analysis and introduces a four-axis framework covering (A) overall execution plan, (B) index structure, (C) next-step control (strategies and triggers), and (D) stop/continue criteria. Using this schema, we map representative multi-hop QA systems and synthesize reported ablations and tendencies on standard benchmarks (e.g., HotpotQA, 2WikiMultiHopQA, MuSiQue), highlighting recurring trade-offs among effectiveness, efficiency, and evidence faithfulness. We conclude with open challenges for retrieval--reasoning agents, including structure-aware planning, transferable control policies, and robust stopping under distribution shift.
\end{abstract}

\section{Introduction}
\label{sec:intro}

Multi-hop question answering (QA) has become a core testbed for machine reasoning. Compared to single-hop QA, where a single passage often suffices, multi-hop benchmarks such as HotpotQA \cite{yang2018hotpotqa}, 2WikiMultiHopQA \cite{ho2020constructing}, and MuSiQue \cite{trivedi2022musique} require systems to combine several pieces of evidence, compare attributes across entities, or follow chains of relations that span multiple documents \cite{shi2024generate}. Many of these datasets annotate supporting sentences and sometimes explicit reasoning paths, making them natural grounds for analyzing how systems actually reason.

Large language models (LLMs) equipped with retrieval have become the dominant approach for such tasks. Retrieval-augmented generation (RAG) couples a retriever with an LLM and has achieved strong performance on knowledge-intensive QA \cite{lewis2020retrieval,gao2023precise}, while more recent agent-style systems model question answering as a sequence of actions over tools and documents \cite{he-etal-2024-webvoyager,kim-etal-2024-rada,liu2025mt2st,liu2025hsgm,liu2025csv}. However, most of this literature focuses on architectural or training innovations and treats \emph{how retrieval and reasoning interact} as an implementation detail. Concretely, the sequence of retrieval calls, intermediate hypotheses, and stopping decisions is rarely made explicit or analyzed systematically. This omission matters because many multi-hop benchmarks are brittle to single-pass retrieval under distractors or long contexts, motivating explicit retrieval--reasoning loops that adapt queries, evidence selection, and stopping over time.

At the algorithmic level, what distinguishes multi-hop QA from single-hop QA is not only the presence of retrieval, but how retrieval and reasoning are coupled over time. These procedural decisions directly shape reported trade-offs in accuracy, efficiency, and robustness, and motivate treating the retrieval--reasoning process as an explicit object of analysis.

% Once made explicit, the same procedure also serves as a natural process-level explanation of the model’s behavior : each retrieval call and each intermediate conclusion document why a particular piece of evidence was used and how the final answer was assembled.

\begin{figure*}[t]
    \centering
    \includegraphics[width=\linewidth]{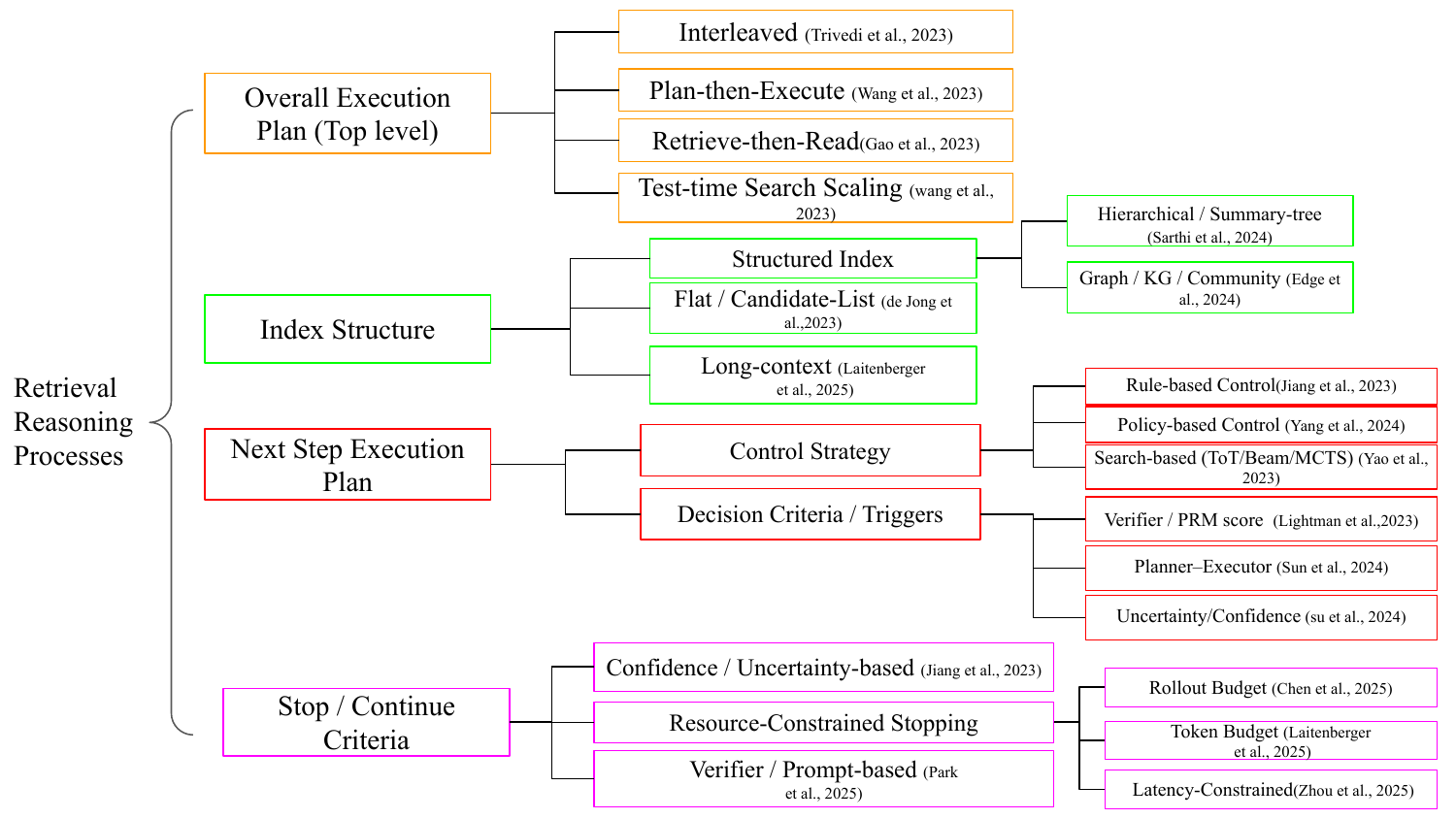}
    \caption{
      Overview of the four design axes we use to describe retrieval--reasoning processes in multi-hop QA:
      (A) overall execution plan, (B) index structure, (C)  next step execution plan, and (D) stop/continue criteria.
      The framework abstracts over concrete model architectures and datasets.
    }
    \label{fig:method-map}
\end{figure*}
Unlike existing surveys that primarily organize methods by modules, stages, or training loops, our framework treats the execution procedure as the unit of analysis. This emphasizes how retrieval, reasoning, control, and stopping decisions are interleaved over time, rather than discussed as separable components. As a result, design choices that are often presented as method-specific heuristics—e.g., controller triggers and stop/continue signals—become alignable across model families under a shared axis-based schema.
We view a retrieval--reasoning process as the execution procedure that interleaves retrieval, state updates, action selection, and stopping; our goal is to make these procedural choices explicit and comparable across systems rather than to claim causal superiority.

We organize the survey around four questions: (RQ1) why make the procedure explicit, (RQ2) what are the key procedural design axes, (RQ3) what trade-offs are reported, and (RQ4) what open challenges remain.
Our paper list is available at \url{https://anonymous.4open.science/r/MultihopQA-Reasoning-Retrieve-Survey-3324/}.

\section{Background}
\label{sec:background}

\subsection{Multi-hop QA Datasets}

Multi-hop question answering (QA) requires a system to combine
information from multiple pieces of evidence to answer a single
question. Datasets such as HotpotQA, 2WikiMultiHopQA, and MuSiQue were explicitly designed to test this ability \cite{yang2018hotpotqa,ho2020constructing,trivedi2022musique}. HotpotQA provides Wikipedia-based questions with sentence-level supporting facts and adversarial distractor paragraphs \cite{yang2018hotpotqa}. 2WikiMultiHopQA mixes structured and unstructured evidence and includes annotated reasoning paths \cite{ho2020constructing}, while MuSiQue composes single-hop questions into 2–4 hop questions and controls for artifacts  and disconnected reasoning \cite{trivedi2022musique}. More recent datasets such as MEQA and FanOutQA further emphasize multi-document aggregation and the completeness and consistency of supporting evidence \cite{li2024meqa,zhu2024fanoutqa,cai2025does}. 

\subsection{Existing Surveys and Remaining Gaps}
Several surveys and tutorials organize the rapidly growing literature on multi-hop QA and reasoning-capable LLMs. Mavi and Jangra \cite{mavi2024multi} review multi-hop QA datasets and model architectures, focusing on how graphs, question decomposition, and memory mechanisms support complex reasoning. Other surveys discuss reasoning LLMs more broadly, covering prompting strategies, training recipes, and cognitive perspectives \cite{li2025system,sui2025stop,niu2025decoding}, or analyse when parametric knowledge needs to be complemented by external tools and retrieval \cite{li2025knowledge}. These overviews cover many \emph{ingredients} relevant to our focus (decomposition, retrieval modules, memory, verification), but they typically organize the space around \emph{components}, stages, or model families (e.g., graph-based vs.\ decomposition-based; RAG vs.\ non-RAG) rather than treating the \emph{execution procedure}---the step-by-step coupling between retrieval and reasoning---as the primary unit of analysis \cite{gao2023retrieval}. Consequently, procedural questions such as \emph{when retrieval is invoked relative to reasoning}, \emph{how evidence is organized and retained under context budgets}, \emph{how the next action is chosen}, and \emph{how stop/continue decisions are made} often appear only implicitly as method-specific design details or scattered heuristics. In particular, across the above surveys, stop/continue is rarely separated as a first-class procedural design choice, and the literature is not typically mapped into an execution-oriented coding schema that would enable aligned, reproducible comparisons across model families. In contrast, our framework takes the execution procedure itself as the unit of analysis and makes retrieval, reasoning, control, and stopping decisions explicitly comparable along shared axes.

% multi-hop QA and reasoning-capable LLMs. Mavi and Jangra \cite{mavi2024multi} review multi-hop QA datasets and model architectures, focusing on how graphs, question decomposition, and memory mechanisms support complex reasoning. Other surveys discuss reasoning LLMs more broadly, covering prompting strategies, training recipes, and cognitive perspectives \cite{li2025system,sui2025stop}, or analyse when parametric knowledge needs to be complemented by external tools and retrieval \cite{li2025knowledge}.

\section{Why an Explicit Retrieval--Reasoning Process?}
\label{sec:rq1}
Our first research question (\textbf{RQ1}) asks why it is useful to treat multi-hop QA as an explicit retrieval--reasoning \emph{procedure} rather than as a one time retrieve and read.  The core reason is that, for knowledge-intensive multi-step questions, the \emph{procedure}---when to retrieve, how to form the next query, what evidence to keep under budgets, and when to stop---is often the main lever that changes both accuracy and failure modes. For example, IRCoT improves multi-step QA by \emph{interleaving} retrieval with chain-of-thought, i.e., by changing the retrieval--reasoning interaction rather than only scaling the reader \cite{trivedi2023interleaving}. Moreover, agentic QA methods such as ReAct make tool calls and evidence usage explicit, turning the trajectory itself into a process-level account of how an answer is obtained \cite{yao2022react}. Finally, recent work treats stopping as a first-class decision (instead of a fixed hop budget), highlighting that \emph{continue vs.\ stop} is part of the procedure and directly controls cost and robustness \cite{park2025stop}.

% Our coding of 252 recent multi-hop QA/RAG papers (App.~\ref{sec:prisma-oldtopic}) further suggests that many recent advances can be reinterpreted as innovations in retrieval--reasoning coupling, motivating a survey that organizes the field around procedural design choices and their trade-offs.

% Based on our coding of 252 recent multi-hop QA and RAG papers (App.~\ref{sec:prisma-oldtopic}), the topical dimension we call ``retrieval--reasoning coupling'' is where most methods actually innovate: 104 of 252 papers were rated as high-fit on this dimension. Typical examples include interleaving chain-of-thought with retrieval \cite{trivedi2023interleaving}, learning multi-hop retrieval beams \cite{zhang2024end}, using hierarchical or graph-structured indices with custom controllers \cite{sarthi2024raptor,edge2024local}, and adding sufficiency-aware critics in multi-round RAG \cite{xu-etal-2025-simrag,ji-etal-2025-resource}.

% By contrast, purely retrieve-then-read baselines that retrieve a large context once and let a long-context LLM answer directly \cite{jiang2024longrag,zhao2024longrag} struggle on adversarial multi-hop benchmarks such as HotpotQA and MuSiQue, where distractor passages can cause the model to attend to irrelevant but lexically similar content \cite{yang2018hotpotqa,trivedi2022musique} and provide little visibility into which sub-questions were solved in which order.

\section{Design axes of Retrieval--Reasoning Procedures}
\label{sec:framework}

To answer \textbf{RQ2}, we describe multi-hop QA systems in terms of four
design axes Figure~\ref{fig:method-map} Each axis abstracts over many concrete modules and implementations; our goal is to provide a shared vocabulary for characterizing execution procedures.

\begin{figure*}[t]
    \centering
    \includegraphics[width=\linewidth]{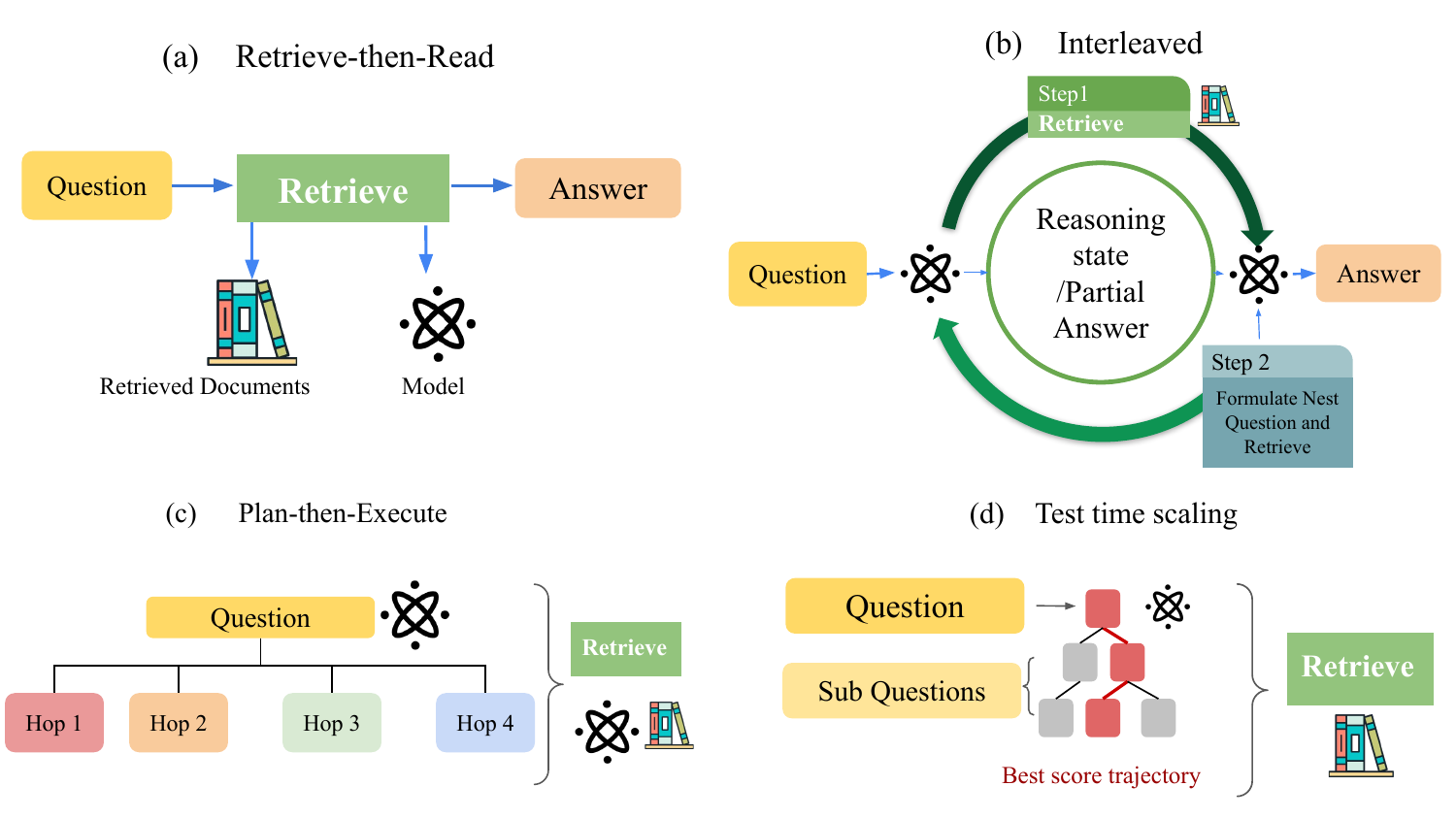}
    \caption{
     overall execution plans for retrieval--reasoning in multi-hop QA (Axis~A of our framework).
(a) \emph{Retrieve--then--Read}: a single retrieval pass feeds the reader model.
(b) \emph{Interleaved}: the model alternates reasoning with additional retrieval calls, updating a partial answer/state.
(c) \emph{Plan--then--Execute}: the model first decomposes the question into sub-hops, then retrieves and answers for each hop following the plan.
(d) \emph{Test-time search scaling}: the model explores multiple candidate reasoning trajectories over the question/sub-questions and selects the best-scoring path for retrieval and answering.}
    \label{fig:overall-plan}
\end{figure*}

\subsection{Overall Execution Plan}

The overall execution plan specifies \emph{when} retrieval is invoked relative to reasoning.  We highlight four recurring patterns that correspond to the sub-panels in Figure~\ref{fig:overall-plan}: retrieve–then–read, interleaved retrieval, plan–then–execute, and test-time search scaling.

\textbf{Retrieve–then–Read.}
Classical retrieval-augmented QA pipelines retrieve a fixed set of passages given the original question, then feed them once into a reader or generator.  Examples include DrQA-style open-domain \cite{chen2017reading}, DPR + BERT readers \cite{karpukhin2020dense}, FiD-style fusion-in-decoder models \cite{izacard2021leveraging}, and the original RAG model \cite{lewis2020retrieval} for knowledge-intensive QA. Multi-hop extensions such as MDR  \cite{qi-etal-2019-answering} still follow the same pattern: the retriever is called a small fixed number of times, and the reader operates in a single, long-context pass.

\textbf{Interleaved Retrieval and Reasoning.}
Interleaved plans alternate between reasoning steps and retrieval calls. Self-Ask \cite{press-etal-2023-measuring} and ReAct \cite{yao2022react} are early examples that prompt the LLM to produce intermediate thoughts and explicit \textsc{Search}/\textsc{Lookup} actions.  IRCoT \cite{trivedi2023interleaving} couples chain-of-thought with retrieval on HotpotQA and MuSiQue.  FLARE \cite{jiang2023active} and DRAGIN  \cite{su-etal-2024-dragin} further refine when and how new queries are issued during generation.  Lost-in-Retrieval analyzes the failure modes of such loops and proposes targeted query reformulation and filtering for multi-hop RAG.
% In our framework, all of these systems implement an \emph{interleaved}
% overall execution plan where retrieval and reasoning co-evolve over time.
% In our framework, all of these systems instantiate an interleaved overall execution plan.

\textbf{Plan–then–Execute.}
Plan–then–execute methods first produce an explicit reasoning or decomposition plan, and only then execute retrieval and answering steps according to that plan.  Decomposed Prompting \cite{khot2022decomposed}, Plan-and-Solve \cite{wang-etal-2023-plan}, and Least-to-Most Prompting \cite{zhou2022least} all decompose questions into sub-questions that can be answered sequentially.  QDMR \cite{wolfson-etal-2020-break} and recent QDMR-based multi-hop QA systems use structured decomposition graphs as plans.  In retrieval settings, PAR-RAG \cite{zhang2025credible} and BEAM Retrieval \cite{zhang2024end} first plan candidate evidence chains and then execute retrieval over them, sometimes with verifier feedback.

\textbf{Test-Time Search Scaling.}
A more aggressive family of overall execution plans treats the whole retrieval–reasoning process as a search problem and scales performance by exploring many candidate trajectories at test time.  Tree-of-Thoughts \cite{yao2023tree} and Graph-of-Thoughts \cite{besta2024graph} use explicit search over thought trees and graphs. MindStar \cite{kang2024mindstar}, MCTS-DPO \cite{xie2024monte}, and Monte-Carlo-tree self-refinement methods apply MCTS-style search guided by value functions or process reward models  like ReARTeR \cite{sun2025rearter}.  In knowledge-intensive settings, RAP \cite{hao2023reasoning} plans over world models, and several GraphRAG \cite{han2024retrieval} variants explore multiple graph traversal paths before committing to an answer.  
\begin{figure}[t]
    \centering
    % Replace this with your actual figure file
    \includegraphics[width=\linewidth]{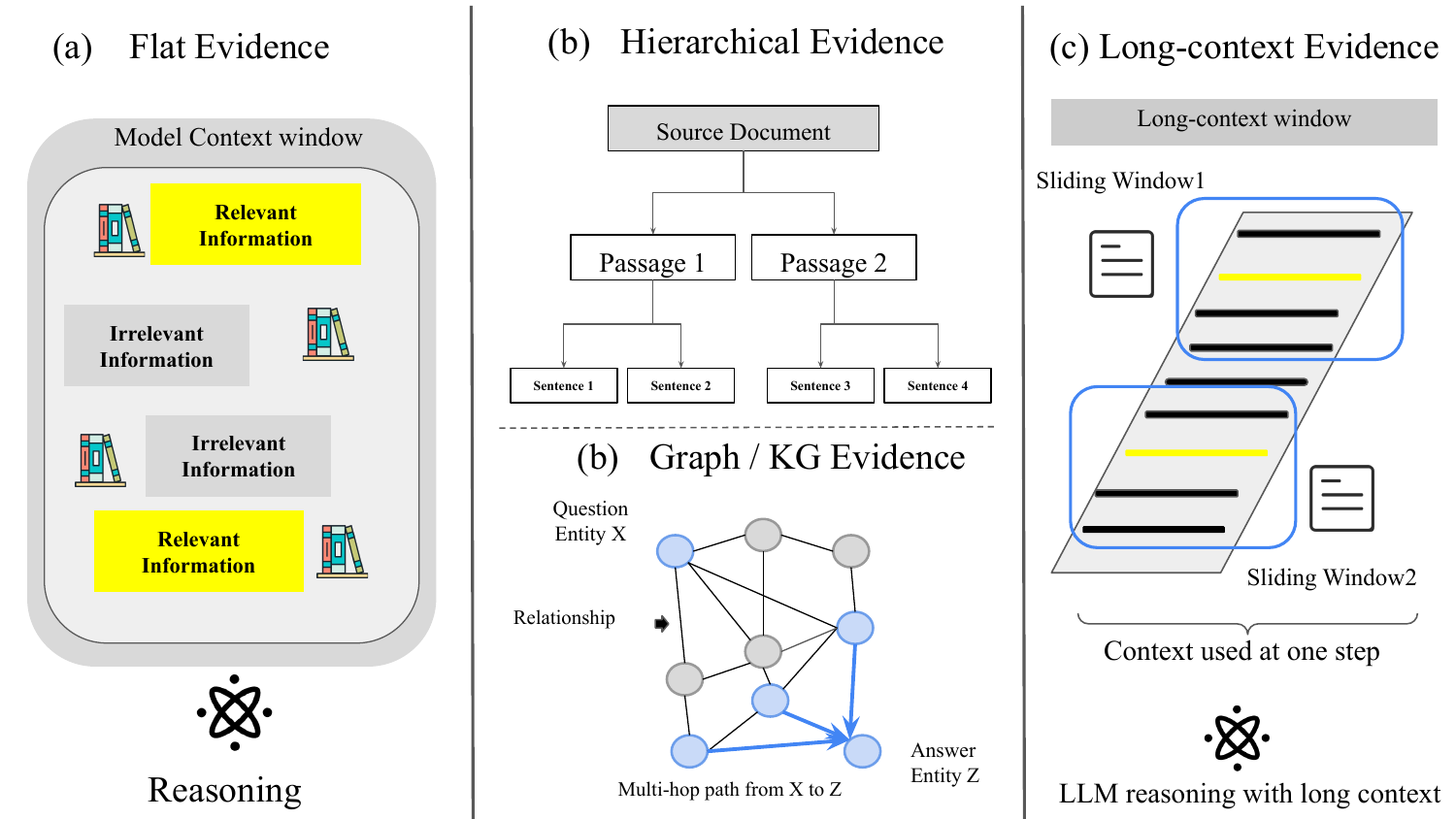}
    \caption{ Contrasts flat passage lists, hierarchical / tree-structured indices, graph-based indices, and long-context storage.
    }
    \label{fig:index_structure}
\end{figure}

\subsection{Index Structure}
\label{sec:index}

Index structure specifies \emph{how} external knowledge is organized and accessed.  Figure~\ref{fig:index_structure} distinguishes four
common forms: flat candidate lists, hierarchical or summary trees, graph- or KG-based structures, and long-context evidence units.

\textbf{Flat / Candidate-List Indices.}
The default in most multi-hop QA systems is a flat collection of passages or documents indexed by BM25 or dense embeddings.  DrQA \cite{chen2017reading}, DPR \cite{karpukhin2020dense}, FiD \cite{izacard2021leveraging} and RAG all adopt this structure in open-domain QA, Multi-hop retrievers such as MDR \cite{xiong2020answering}, HopRetriever \cite{li2021hopretriever}, and step-wise dense retrievers for HotpotQA and 2WikiMultiHopQA  still operate over flat indices, even when they train separate retrievers per hop.  
\begin{table}[t]
\centering
\scriptsize
\setlength{\tabcolsep}{3pt}
\renewcommand{\arraystretch}{1.05}
\begin{tabularx}{\columnwidth}{@{}lccccX@{}}
% \begin{tabularx}{\columnwidth}{@{}p{2.2cm}ccccX@{}}
\toprule
\textbf{System} & \textbf{A} & \textbf{B} & \textbf{C} & \textbf{D} & \textbf{Benchmarks (typical)} \\
\midrule
DrQA / DPR+FiD & R & F & R & B & NQ, TriviaQA \cite{chen2017reading,karpukhin2020dense,izacard2021leveraging} \\
IRCoT & I & F & R & B & HotpotQA, MuSiQue, IIRC \cite{trivedi2023interleaving,trivedi2022musique} \\
DEC & I & F & P & B/H & HotpotQA \cite{ji-etal-2025-resource} \\
Beam Retrieval & P & F & P & B & HotpotQA, 2Wiki, MuSiQue \cite{zhang2024end} \\
PAR-RAG & P & F/G & V & B & HotpotQA, 2Wiki \cite{zhang2025credible} \\
RAPTOR & R/I & H & R & B & Long-doc QA, Hotpot variants \cite{sarthi2024raptor} \\
LongRAG & R/I & L & R & B/H & HotpotQA, ODQA \cite{jiang2024longrag,zhao2024longrag} \\
GraphRAG & I & G & R/P & B & Sensemaking, MHQA \cite{edge2024local} \\
KG-o1 & P/I & G & P/V & B & 2Wiki, KGQA \cite{wang2025kg} \\
SIM-RAG & I & F & P/V & C/L & Multi-round RAG \cite{xu-etal-2025-simrag} \\
Stop-RAG & I/P & F & P & L & Iterative RAG / MHQA \cite{park2025stop} \\
\bottomrule
\end{tabularx}
\caption{
 Representative systems mapped into our four-axis framework.
A: overall execution plan (R=retrieve--then--read, I=interleaved, P=plan--then--execute, S=search scaling).
B: index structure (F=flat, H=hierarchical, G=graph/KG, L=long-context).
C: control (R=rules, P=learned policy, V=verifier-gated, S=search-based).
D: stopping (B=budget, C=confidence-based, L=learned/value-based).}
\label{tab:systems-matrix}
\end{table}

\textbf{Hierarchical / Summary-Tree Indices.}
Hierarchical indices organize documents into multi-level clusters or summary trees, allowing systems to retrieve coarse summaries first and zoom in on relevant subtrees.  RAPTOR \cite{sarthi2024raptor} builds recursive summary trees and shows that retrieving via tree nodes can substantially reduce tokens while preserving answer accuracy.  Other long-document QA systems build section-based or chapter-based hierarchies and retrieve using both coarse and fine nodes \cite{xia2025selection}.  LongRAG \cite{zhao2024longrag} uses long retrieval units (e.g., 4K tokens) as coarse segments and lets the LLM decide which parts to attend to, effectively creating a two-level index structure.  Hierarchical community-based indices in GraphRAG also fall into this category when communities are organized in a tree \cite{edge2024local}.

\textbf{Graph / KG / Community Structures.}
Graph-based structures connect entities, documents or passages via explicit edges, enabling path-based reasoning and community-level summarisation.  Early multi-hop QA work over KGs includes GraftNet \cite{liu2022graftnet}, PullNet \cite{sun2019pullnet}, and related graph neural architectures.  More recent GraphRAG systems construct entity and document graphs from text and perform retrieval along graph edges or communities \cite{edge2024local}.  KG-augmented RAG approaches such as KG-FiD \cite{yu2022kg}, UniKGQA \cite{jiang2022unikgqa} and KG-o1 \cite{wang2025kg} extend this idea by combining KG traversal with text retrieval for multi-hop QA. 
% These systems instantiate a graph- or community-structured retrieval backbone.

\textbf{Long-Context Evidence.}
Instead of index, some methods rely on long-context LLMs and store very long segments or entire documents as retrieval units. LongRAG explicitly proposes long units to reduce the number of hops needed to cover the required evidence \cite{zhao2024longrag,lee-etal-2025-loft,das2025can}.  LongT5 and other long-context similarly retrieve and process long sequences in a  retrieve and read  \cite{guo2022longt5,li2024long}.

\subsection{Next-Step Execute Plan}
\label{sec:control}

The \emph{next-step execute plan} specifies how a system chooses its next action---retrieve again, expand candidates, verify, backtrack, or answer---given the current state. 
% We separate this component into (1) a \textbf{control strategy}, which proposes candidate actions, and (2) \textbf{decision criteria / triggers}, which determine whether to continue, stop, or revise.  
We separate this component into (1) a control strategy, which proposes candidate actions, and (2) decision criteria / triggers, which determine whether to continue expanding or to revise/hand off control (termination is covered in Axis D).

% \paragraph{Control strategy.}

\textbf{Rule-based control.}
Many systems instead hard-code a reasoning control pattern: a fixed alternation of \textsc{Reason} and \textsc{Search} steps, simple heuristics for query reformulation, and manually tuned filters over retrieved passages. Self-Ask, ReAct, IRCoT and DEC all fall in this family, using prompts and hand-designed rules to decide when to ask a follow-up question, when to call the retriever, and when to output an answer \citep{press-etal-2023-measuring,yao2022react,trivedi2023interleaving,ji-etal-2025-resource}. Rule-based controllers are easy to implement and robust to small data regimes, but they do not learn from failures and often need re-tuning when the domain or retriever changes.

\textbf{Policy-based control.}
Here the next action is chosen by a learned policy that observes the question, retrieved evidence and partial reasoning trace. Examples include beam-style multi-hop retrievers with learned expansion policies \citep{zhang2024end}, RAG agents trained to select between tools or retrieval templates via supervision or reinforcement learning \citep{yang2024rag,park2025stop}, and multi-round RAG systems whose controllers are fine-tuned on inner-monologue trajectories \citep{xu-etal-2025-simrag}.

\textbf{Search-based control.}
A third typical treats next-step selection as search over partial trajectories.  Tree-of-Thoughts explores a tree of reasoning states and uses value or vote-based scores to decide which branches to expand \citep{yao2023tree}.  MindStar and related MCTS-style methods perform Monte-Carlo search over tool-use and reasoning sequences, guided by process reward models or learned value functions \citep{kang2024mindstar,xie2024monte}.  Beam-style search over plans or KG paths similarly expands multiple candidate chains in parallel before committing to an answer \citep{zhang2024end,jiang2022unikgqa}. These approaches can substantially improve long-horizon success, but incur higher inference cost and rely on strong scoring signals.

% \paragraph{Decision criteria / triggers.}
% Decision criteria determine when the controller should continue expanding, hand control to another module, or stop.

\textbf{Verifier / PRM-score triggers.}
Verifier-gated loops insert an explicit judge---an NLI-style checker, a process reward model, or a self-evaluation module---and use its score as a trigger.  Process reward models that score intermediate reasoning steps \citep{lightman2023let} and hallucination detectors such as SelfCheckGPT \citep{manakul2023selfcheckgpt} exemplify this pattern: low-scoring answers or chains trigger extra retrieval or revision instead of termination.  Tool-centric systems like TaskMatrix-style agents also rely on verifier-like scoring over tool outputs to decide whether a subgoal is solved or whether another tool call is needed \citep{liang2024taskmatrix}.  In RAG settings, verifier / PRM triggers are often combined with sufficiency critics, as in SIM-RAG and related multi-round controllers \citep{xu-etal-2025-simrag,zhang2025credible}.

\textbf{Planner--executor triggers.}
Some agents separate planning and execution and use the plan’s status as an explicit trigger.  Planner–executor architectures first generate a structured plan (sub-questions, tool calls, or graph operations) and then execute it step-wise, revising or extending the plan when execution fails or when critical preconditions are not met \citep{khot2022decomposed,wang2025pecan,sun2025rearter}.  In this view, the decision criterion is whether the current plan still appears feasible and sufficiently specified; negative feedback triggers re-planning or refinement, while successful plan completion triggers termination.  This pattern is especially common in web and tool agents that must coordinate several API calls \cite{he-etal-2024-webvoyager, chen2025lightweight,li2025bideeplab}.

\textbf{Uncertainty / confidence triggers.}
Finally, uncertainty-aware triggers estimate how likely the current answer (or intermediate state) is to be correct or sufficiently supported, and continue retrieving or reasoning only while confidence is below a threshold.  Signals include calibrated answer probabilities, disagreement across sampled chains, or explicit ``uncertainty of thoughts'' scores \citep{hu2024uncertainty,sui2025stop}.  Dynamic context-cutoff policies similarly stop reading long inputs once new tokens are unlikely to change the prediction \citep{xie2025knowing}. In multi-round RAG, such confidence signals are often combined with learned sufficiency critics or value-based controllers \citep{xu-etal-2025-simrag,park2025stop}, yielding adaptive-depth retrieval loops that trade a small amount of computation for more stable behavior under distribution shift.

\subsection{Stop / Continue Criteria}
\label{sec:stop}
Stop / continue criteria specify \emph{when} the retrieval--reasoning procedure terminates. In multi-hop QA, stopping too early misses necessary evidence, while stopping too late inflates prompt length and latency and increases exposure to distractors.

\textbf{Resource-constrained stopping.}
Many systems ignore the detailed state and instead impose hard budgets on search. A first subtype is a \emph{rollout budget}: interleaved agents such as Self-Ask, ReAct, IRCoT and DRAGIN restrict the number of reasoning or tool-use steps (e.g., at most $K$ \textsc{Search} calls or $H$ decomposition steps) and must answer once the budget is exhausted \cite{press-etal-2023-measuring,yao2022react,trivedi2023interleaving,su2024dragin}. A second subtype is a \emph{token budget}, where multi-hop RAG pipelines (e.g., LongRAG, DEC and budget-aware CoT (chain of thought) methods) fix the maximum retrieved or prompt tokens per question and truncate extra evidence or thoughts when this limit is reached \cite{jiang2024longrag,ji-etal-2025-resource,han2025your}. A third subtype is
\emph{latency-constrained} stopping, where web or tool agents terminate once a wall-clock or API-call budget is consumed \cite{shen2024taskbench}.

\textbf{Confidence- / uncertainty-based stopping.}
A second family asks how \emph{confident} the model (or a small auxiliary head) is in the current answer or intermediate state, and stops once uncertainty falls below a threshold. Representative signals include logit margins, entropy, and disagreement across self-consistent samples, as surveyed in work on ``stop overthinking'' and early-exit reasoning \cite{sui2025stop,li2025system,11050741}. Uncertainty-of-thought approaches explicitly estimate the uncertainty of candidate thoughts and decide whether to produce more intermediate steps or to answer directly \cite{hu2024uncertainty}, while dynamic context-cutoff methods learn when an LLM has read enough of a long input and can safely stop consuming further tokens \cite{xie2025knowing,11144414,11257590}. 
% Compared to pure budget rules, confidence-based criteria adapt search depth to question difficulty, but their reliability depends on how well the chosen uncertainty proxy is calibrated under distribution shift.

\textbf{Verifier- / prompt-based stopping.}
The third family inserts an explicit \emph{checking} stage and delegates the stop / continue decision to a verifier. Rationale-refinement and fact-checking pipelines such as RARR, SelfCheckGPT and Chain-of-Verification first generate a draft answer and then ask a verifier (often another LLM or an NLI-style model) whether each claim is sufficiently supported by the retrieved evidence; a positive verdict triggers termination, while unsupported claims trigger additional retrieval, clarification questions or answer revision \cite{gao2023rarr,manakul2023selfcheckgpt,dhuliawala-etal-2024-chain}. In multi-round RAG, sufficiency critics and value-based controllers such as SIM-RAG, Stop-RAG, DEC and ReARTeR can be viewed as \emph{learned verifiers}: they inspect the question, retrieved evidence and inner-monologue trace and decide whether another retrieval round is expected to help, using QA accuracy or process rewards as supervision \cite{xu-etal-2025-simrag,park2025stop,ji-etal-2025-resource, sun2025rearter}.
% These verifier- / prompt-based criteria typically improve factuality and evidence faithfulness, but introduce extra verifier calls and therefore trade a small amount of efficiency for more reliable stopping.
Table \ref{tab:systems-matrix} maps representative systems onto these four axes.

\section{RQ3: Reported empirical tendencies in retrieval–reasoning design}
\label{sec:rq3}

This section performs a pattern-mining synthesis of empirical findings as reported in the original papers. Because results are obtained under heterogeneous retrievers, LLMs, prompts, and budgets, we do not claim causal superiority of any axis choice. Instead, we summarize directional tendencies that (1) recur across multiple studies, or (2) are supported by within-paper ablations under matched budgets, and we treat cross-paper comparisons as suggestive rather than definitive.

\textbf{Overall execution plan.}

Compared to retrieve--then--read (single-pass retrieval) baselines, several studies report an association between explicitly multi-step execution plans and higher answer F1 and joint answer--evidence scores, particularly in full-wiki and distractor-heavy settings \citep{trivedi2023interleaving}.
Several studies report an association between explicitly multi-step execution plans and higher
Interleaved plans that alternate chain-of-thought and retrieval, such as IRCoT-style interleaving is reported to increase recall and joint F1 in on HotpotQA, 2WikiMultiHopQA and MuSiQue \citep{trivedi2023interleaving}.  Within the reported settings (often under matched budgets within the same study), plan–then–execute methods are reported to reach comparable gains while batching sub-queries \citep{zhang2024end,zhang2025credible}.
Test-time search-scaling methods (e.g., Tree-of-Thoughts, MindStar, MCTS-style search) further boost long-horizon accuracy by exploring many candidate trajectories, but incur the highest latency and compute costs \citep{yao2023tree,kang2024mindstar,xie2024monte}.

\textbf{Index structure.}
Most multi-hop QA systems still operate over flat BM25 or dense passage collections, and attribute gains mainly to changes in overall execution plan or control policy rather than radically different indices \citep{chen2017reading,karpukhin2020dense,trivedi2023interleaving,zhang2024end,ji-etal-2025-resource}. Index structures primarily are discussed in relation to efficiency and path faithfulness. Hierarchical or summary-tree structures such as RAPTOR reduce context length on long-document QA by retrieving summaries and then drilling down to raw passages, typically preserving or slightly improving EM \citep{sarthi2024raptor}.
% Long-unit / long-context designs like LongRAG retrieve fewer but longer segments and, when paired with long-context LLMs, match or surpass trained multi-hop baselines on HotpotQA and open-domain QA without additional finetuning \citep{jiang2024longrag,zhao2024longrag}.
Graph- and KG-based index structures (e.g., GraphRAG, KG-o1) organize entities and documents into graphs or communities and retrieve along paths, improving joint answer+path metrics and producing more coherent evidence chains, at the cost of graph construction and maintenance \citep{edge2024local,wang2025kg}.

\textbf{From control and triggers (Axis C/D) to outcomes.}
For the next-step execute plan, our coded systems are consistent with the hypothesis that three robust patterns. First, moving from purely rule-based control with static budgets to learned or planner-style control tends to improve answer accuracy at similar average cost.

Beam Retrieval and PAR-RAG then replace hand-written rhythms with plan-driven controllers and beam-style expansion: Beam Retrieval reaches 69.2 answer F1 / 91.4 support F1 on MuSiQue, and PAR-RAG reports 2Wiki and HotpotQA EM gains of roughly 15--30 points over their strongest non-planning baselines under matched or smaller budgets \cite{zhang2024end,zhang2025credible}. These results suggest that global, plan-aware controllers are especially beneficial when the evidence graph is rich and hop depth varies widely.

Second, combining policy-based control with verifier or PRM-style triggers are repeatedly reported to coincide with improved faithfulness and joint answer+evidence scores. Verifier-gated refinement in RARR and Chain-of-Verification already reduces unsupported claims compared to single-pass RAG and vanilla CoT \cite{gao2023rarr,dhuliawala-etal-2024-chain}. 
% PAR-RAG extends this pattern to multi-hop RAG: a verifier scores candidate plans and executions, yielding 7--12 F1 improvements over IRCoT+HippoRAG on HotpotQA and 2Wiki while keeping the average number of retrieval steps within the same order of magnitude \cite{zhang2025credible}. 
Similar trends appear in search-based controllers such as MindStar and MCTS-DPO, where PRM-guided tree search prunes low-scoring trajectories and improves math or strategy benchmarks at the cost of a small constant factor in inference time \cite{kang2024mindstar,xie2024monte}. In our framework,  ``policy/search + verifier/PRM'' are  effective on noisy-retrieval or long-horizon tasks where hallucinated chains are common.
Third, uncertainty- or value-based stopping (Axis D) helps controllers avoid both over- and under-search and stabilizes latency under shift\cite{xu-etal-2025-simrag,park2025stop}. Overall, the most robust designs in our corpus pair an adaptive controller (policy- or search-aware, Axis C) with verifier/PRM- or sufficiency-based triggers (Axis D), especially under noisy retrieval and long horizons. Static hop/token budgets can be competitive when tuned in-domain, but they transfer poorly across datasets with different hop and distractor profiles.

\section{RQ4: Open Problems and Future Directions}
\label{sec:open-problems}

\textbf{Challenge 1: Aligning overall execution plans with index structures.}
Most systems fix an \emph{index structure} (flat passages, hierarchical summaries, graphs, or long-context units) and then adopt an \emph{overall execution plan} (retrieve--then--read, interleaved, plan--then--execute, or test-time search scaling) in an ad hoc manner. For example, IRCoT, Beam Retrieval, and DEC operate over flat indices and primarily innovate in scheduling and control  \citep{trivedi2023interleaving,zhang2024end,ji-etal-2025-resource},  whereas RAPTOR, LongRAG, and GraphRAG redesign the structure but often retain standard multi-hop schedules \citep{sarthi2024raptor,jiang2024longrag,edge2024local}. A principled mapping between structure and plan is still missing.
Promising directions include factorial sweeps over the overall execution plan and index strutures under matched budgets, and controllers that choose among multiple (plan, structure) configurations based on query difficulty and resource constraints.

\textbf{Challenge 2: Scalable and adaptive structured indices.}
Structured indices---such as summary trees (RAPTOR), long units (LongRAG), and graphs (GraphRAG, KG-o1)---can improve efficiency and faithfulness, but they are costly to build and maintain \citep{sarthi2024raptor,jiang2024longrag,edge2024local,wang2025kg}. Most work assumes static corpora and relies on hand-tuned hyperparameters for granularity (e.g., cluster size, unit length, and graph pruning). Future research is needed on self-supervised or task-aware structure induction that optimises these choices for downstream multi-hop objectives, as well as incremental maintenance algorithms that keep trees and graphs up to date as documents evolve.

% \textbf{Challenge 3: Generalisable control policies and decision criteria.}
% Current learned or search-based control policies and verifier/PRM-gated decision rules, as explored in recent work on planning and search-based RAG, are often trained on specific datasets and retrievers and may not generalize across hop distributions, index structures, or domains \citep{zhang2024end,ji-etal-2025-resource,yao2023tree}.
% This lack of transferability undermines the promise of adaptive retrieval--reasoning procedures beyond benchmark-specific tuning.
% A promising direction is to formalize retrieval--reasoning as a process-level decision problem and train controllers with objectives that combine evidence quality, trajectory coherence, for example via preference learning or hybrid reinforcement learning over reasoning trajectories.
\textbf{Challenge 3: Generalisable control policies and decision criteria.}
Current learned or search-based control policies and verifier/PRM-gated decision rules are often trained on specific datasets and retrievers, and may not generalize across hop distributions, index structures, or domains \citep{zhang2024end,ji-etal-2025-resource,yao2023tree}.
This brittleness limits the practical promise of adaptive retrieval--reasoning beyond benchmark-specific tuning. A promising direction is to cast retrieval--reasoning as a process-level decision problem and train controllers with objectives that jointly reward evidence quality, trajectory coherence, and cost, e.g., via preference learning over trajectories or hybrid RL.

% \textbf{Challenge 4: Robust stop/continue criteria under budgets and shift.}
% Most systems still rely on static hop, token, or latency budgets, while adaptive stopping based on confidence, sufficiency, or value estimates has only been evaluated on a limited set of benchmarks and remains poorly calibrated under distribution shift \citep{xu-etal-2025-simrag,park2025stop}.
% This is problematic because over- or under-search directly affects both reliability and efficiency in real-world deployments.
% An important open problem is to develop stop/continue criteria that remain calibrated across retrievers, corpora, and LLM backbones, and to evaluate them under controlled variations of hop depth and retrieval noise.
\textbf{Challenge 4: Robust stop/continue criteria under budgets and shift.}
Most systems still rely on static hop, token, or latency budgets, while adaptive stopping based on confidence, sufficiency, or value estimates is evaluated on limited benchmarks and remains poorly calibrated under distribution shift \citep{xu-etal-2025-simrag,park2025stop}.
This matters because over- and under-search simultaneously degrade reliability and efficiency in deployment.
An important open problem is to develop stop/continue criteria that remain calibrated across retrievers, corpora, and LLM backbones, and to evaluate them under controlled variations of hop depth and retrieval noise.

\section{Conclusion}
We have surveyed retrieval--reasoning in multi-hop question answering through the lens of four research questions. We argued that multi-hop QA requires an explicit retrieval--reasoning process (RQ1), introduced a four-axis design space for describing this process (RQ2), and analyzed how axis-level design choices affect effectiveness, efficiency, and faithfulness on standard benchmarks (RQ3). Finally, we highlighted several open problems in designing and evaluating retrieval--reasoning systems (RQ4). We hope that our framework helps researchers organize existing methods, design new retrieval reasoning agents in a more principled way.

\section*{Limitations}

First, while we strive to cover a broad range of multi-hop QA systems, the field is evolving rapidly and our selection is necessarily incomplete. Second, we focus primarily on English, text-based benchmarks and do not systematically cover multilingual or multi-modal settings. Third, our empirical synthesis relies on numbers reported in original papers, which may be influenced by differences in hyperparameters, compute budgets, or evaluation scripts. Despite these limitations, we believe the four-axis framework provides a useful starting point for analysing retrieval--reasoning systems.

\bibliography{custom}

\appendix

\section{PRISMA-style Literature Collection}
\label{sec:prisma}

To repeatedly reported identify existing surveys and overviews related to
multi-hop question answering (QA), we followed a PRISMA-style (Preferred Reporting Items for Systematic Reviews and Meta-Analyses) procedure adapted to the NLP setting \cite{page2021prisma}.

\paragraph{Scope and time window.}
Our goal was to collect survey- or review-type papers that discuss multi-hop QA, multi-step or compositional QA, or widely used multi-hop benchmarks (e.g., HotpotQA, 2WikiMultiHopQA, MuSiQue). We restricted our search to publications from 2022 onwards, so as to focus on work in the LLM era and recent retrieval-augmented systems.

\paragraph{Data sources.}
We queried multiple scholarly search engines and digital libraries, including Google Scholar (using \texttt{allintitle} and keyword queries), the ACL Anthology, arXiv, and conference proceedings accessible through ACM and IEEE portals. When a candidate paper appeared in multiple sources, we treated it as a single record.

\paragraph{Search queries.}
For each source, we issued the following keyword combinations (illustrated here in a Google Scholar--style syntax), always with the time filter set to \emph{since 2022}:

\begin{itemize}
  \item \texttt{allintitle: (survey OR review OR taxonomy OR "systematic review" OR tutorial) ("multi-hop question answering" OR multihop OR "multi hop")}
  \item \texttt{("multi-hop question answering" OR multihop OR "multi hop" OR "multi-step reasoning" OR "complex question answering" OR "compositional question answering") (survey OR review OR taxonomy OR "systematic review" OR tutorial) (NLP OR "natural language processing")}
  \item \texttt{("HotpotQA" OR "2WikiMultiHopQA" OR "MuSiQue" OR "IIRC" OR "ComplexWebQuestions" OR "QAngaroo" OR "MedHopQA") (survey OR review OR taxonomy OR tutorial OR overview)}
\end{itemize}

These queries were chosen to (1) capture surveys explicitly labelled as such in the title, (2) retrieve broader surveys and tutorials that cover multi-hop or complex QA within NLP, and (3) find overview papers centered around standard multi-hop benchmarks.

\paragraph{Screening and eligibility.}
We first merged results from all queries and removed exact duplicates
based on title and venue.
We then screened titles and abstracts to exclude papers that were not
surveys, reviews, taxonomies, tutorials, or overviews (e.g., individual
method papers, purely theoretical work, or non-NLP applications).
In a second pass, we read introductions and methodology sections of the
remaining papers to check that multi-hop or complex QA was a primary
focus rather than a minor side topic.
We retained only English-language papers that discussed multi-hop QA,
multi-step/compositional QA, or the above benchmarks in sufficient
detail to be useful for positioning our work.

\paragraph{Data extraction and use in this survey.}
For each included paper we recorded meta-data such as publication year,
venue, target tasks (multi-hop QA, reasoning LLMs, RAG, benchmarks),
and the main organisational axes used by the authors (e.g., model
architecture, prompting strategy, training regime).
These annotations were used in our \emph{Background and Related Work}
section to (1) situate our four-component design framework relative to
existing taxonomies and (2) ensure that our discussion of multi-hop QA
and reasoning surveys is comprehensive with respect to the 2022--2025
literature.

Beyond this PRISMA-style survey collection, individual method papers
(e.g., specific retrieval--reasoning systems or search algorithms) were
added through targeted searches, forward and backward citation tracing,
and domain expertise.
\section{Paper Coding and Topical Distribution}
\label{sec:prisma-oldtopic}

For the analysis in Section~\ref{sec:rq1} we coded each retrieved
multi-hop QA / RAG paper along five topical dimensions:

\begin{itemize}
  \item \textbf{T1: Definitions and boundaries of ``multi-hop''.}
        Papers that explicitly discuss what counts as multi-hop
        reasoning, how many supporting facts or hops are required, or
        how to distinguish genuine multi-hop questions from artifacts .
  \item \textbf{T2: Decomposition strategies.}
        Papers that introduce explicit intermediate representations
        such as chain-of-thought, question decomposition, sub-goal
        planning, programs/graphs/templates, or MCTS-style search.
  \item \textbf{T3: Evidence sufficiency and faithfulness.}
        Papers that focus on guaranteeing sufficient support for answers
        and on aligning reasoning chains with retrieved evidence, often
        via supporting-fact metrics, verification modules or
        LLM-as-judge diagnostics.
  \item \textbf{T4: Retrieval--reasoning coupling.}
        Papers whose main contribution lies in how retrieval and
        reasoning are interleaved or controlled, including interleaved
        CoT--retrieval loops \cite{trivedi2023interleaving}, beam-based
        multi-hop retrievers \cite{zhang2024end}, hierarchical or
        graph-aware retrieval controllers \cite{sarthi2024raptor,edge2024local},
        and sufficiency-aware multi-round RAG
        \cite{xu-etal-2025-simrag,ji-etal-2025-resource}.
  \item \textbf{T5: Evaluation and diagnostics.}
        Papers that propose new diagnostics, process-level metrics, or
        robustness tests for multi-hop reasoning.
\end{itemize}

Among all coded papers, topic~T4 (retrieval--reasoning coupling) constitutes the largest group numbers (139 of 252 papers, 55\%), followed by T2 (decomposition strategies) and T5 (evaluation and diagnostics). This distribution motivates our decision to take the retrieval--reasoning procedure as the primary object of analysis in the main text. 
\label{sec:appendix}
\begin{figure*}[t]
  \centering
  \includegraphics[width=\textwidth]{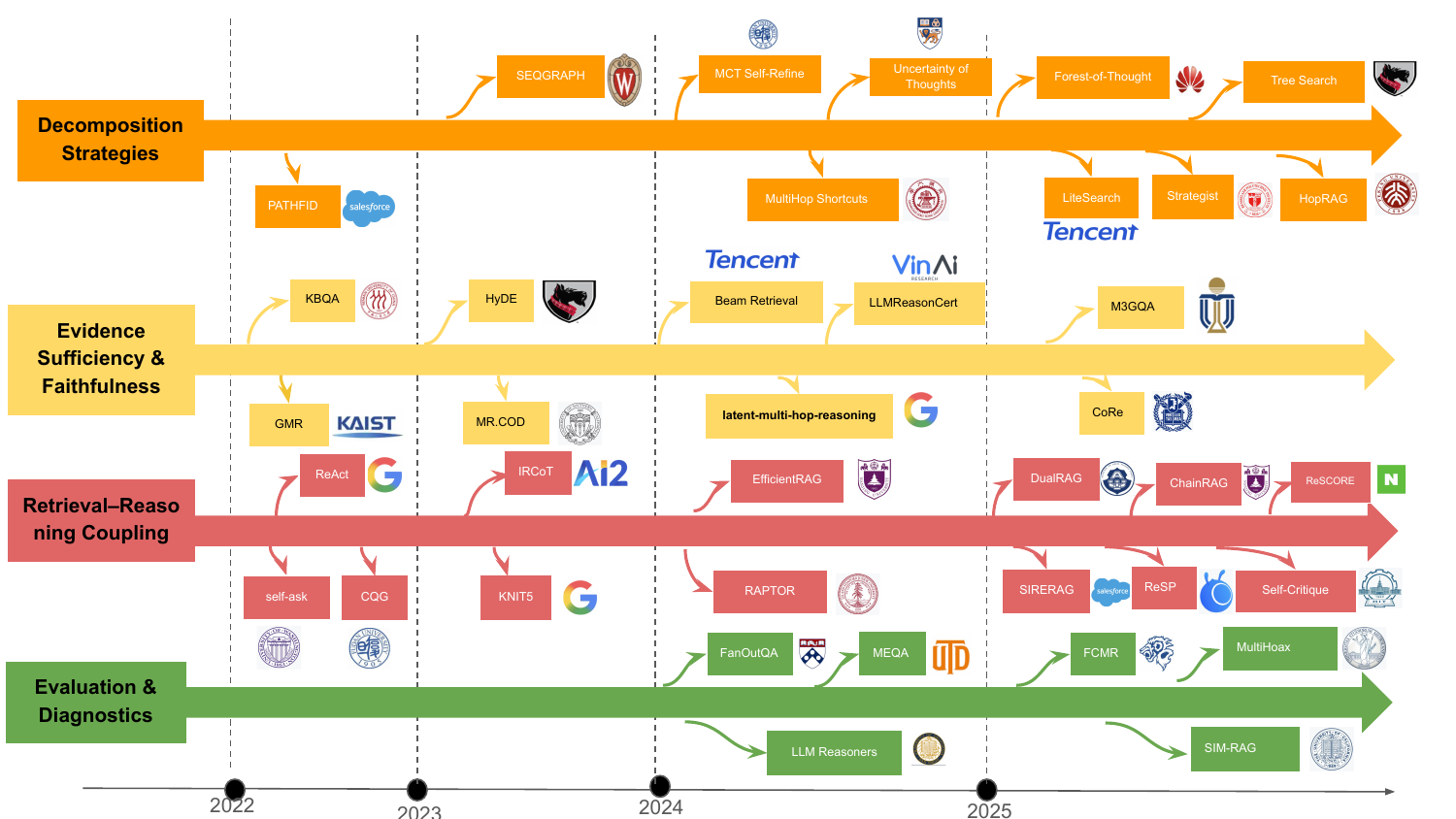}
  \caption{The large topics and how these topics develop  from 2023 to 2025. }

  \label{fig:metrix}
\end{figure*}

% We also have the main company and research institution who write certain paper in Figure \ref{fig:metrix}.
\subsection{Coding Methodology}
\label{sec:coding}

We constructed a paper corpus through keyword-based search (Google Scholar, ACL Anthology, and arXiv) followed by forward/backward citation tracing. After deduplication, we screened titles and abstracts to retain work on multi-hop QA, retrieval-augmented QA, and agentic retrieval--reasoning. For each included paper, we extracted metadata (year, task/benchmarks, supervision, and reported metrics) and coded the method using our four-component schema: overall execution plan, index structure, control policy, and stop/continue criteria. We also assigned each paper to one or more topical dimensions (T1--T5; Appendix~\ref{sec:prisma-oldtopic}). To improve consistency, a subset of papers was double-coded and disagreements were resolved via discussion; the final codes were then applied to the full set. \paragraph{From 252 papers to a 104-paper subset.} 
The full corpus contains 252 multi-hop QA/RAG papers used for the broad topical analysis in Section~\ref{sec:rq1}. For axis-level distributions and method maps, we focus on a 104-paper subset that provides sufficient methodological detail to instantiate our four components (overall execution plan, index structure, control policy, and stop/continue criteria). This subset is therefore a detail-available, framework-compatible slice of the larger corpus rather than a separate collection.

\label{sec:prisma-oldtopic}
\begin{table}[t]
\centering
\small
\begin{tabular}{lcccc}
\hline
\textbf{Topic} & \textbf{High} & \textbf{Med} & \textbf{Low} & \textbf{Total} \\
\hline
T1: Definitions / boundaries      &  6 & 37 & 209 & 252 \\
T2: Decomposition strategies      & 32 & 33 & 187 & 252 \\
T3: Sufficiency / faithfulness    & 17 & 51 & 184 & 252 \\
T4: Retrieval--reasoning coupling & 139 & 14 &  99 & 252 \\
T5: Evaluation / diagnostics      & 29 & 38 & 185 & 252 \\
\hline
\end{tabular}
\caption{
Number of papers coded as High / Medium / Low fit for each topical
dimension.
Retrieval--reasoning coupling (T4) is the largest group in terms of
high-fit papers, with 139 of 252 (about 55\%) methods focusing primarily
on how retrieval and reasoning are coupled.
}
\label{tab:topic-fit}
\end{table}

\begin{table*}[t]
\centering
\small
\setlength{\tabcolsep}{3pt}  % squeeze columns a bit
\begin{tabularx}{\linewidth}{l l c Y}
\toprule
\textbf{Dataset} & \textbf{Type} & \textbf{\#papers} & \textbf{Typical role in Topic~4 papers} \\
\midrule
HotpotQA &
Wiki 2-hop RC (supporting facts) &
29 &
Core multi-hop benchmark for retrieval--reasoning loops and supporting-fact faithfulness. \\

2WikiMultiHopQA &
Hybrid KG+Wiki multi-hop QA &
15 &
Tests cross-document and KG--text reasoning; usually paired with HotpotQA and MuSiQue. \\

WebQuestionsSP &
KB QA over Freebase (1--2 hops) &
14 &
Standard KG-QA baseline; many Topic~4 systems evaluate graph or KG controllers on it. \\

MuSiQue &
Compositional multi-hop QA (Wiki) &
11 &
Harder multi-hop benchmark built by composing single-hop questions; stresses connected reasoning. \\

ComplexWebQuestions (CWQ) &
Web+KB complex QA (search+SPARQL) &
11 &
Open-web setting with complex questions; used to test decomposition and web-search control. \\

MRAMG &
Multimodal retrieval (image+text) &
6 &
Extends retrieval--reasoning from text-only QA to multimodal RAG and agent-style systems. \\

StrategyQA &
Boolean QA with implicit multi-hop &
5 &
Benchmarks control policies for implicit, commonsense multi-hop reasoning and decomposition. \\

MetaQA &
Movie-domain multi-hop KGQA (1--3 hops) &
4 &
Tests KG traversal policies and beam / search-based controllers over large KGs. \\

CronQuestions &
Temporal KGQA (time-aware) &
4 &
Evaluates controllers that must handle temporal constraints on KG reasoning paths. \\

HybridQA &
Table+text hybrid multi-hop QA &
4 &
Benchmarks retrieval--reasoning over structured tables plus linked Wikipedia passages. \\
\bottomrule
\end{tabularx}
\caption{
Most commonly used datasets in our Topic~4 (retrieval--reasoning coupling) sample.
``\#papers'' counts how many of the 104 coded papers mention each dataset.
}
\label{tab:t4-datasets-main}
\end{table*}

\begin{table}[t]
\centering
\small
\setlength{\tabcolsep}{4pt}
\begin{tabularx}{\columnwidth}{Y Y r}
\toprule
\textbf{Dataset} & \textbf{Type} & \textbf{\#papers} \\
\midrule
TriviaQA &
Open-domain QA (Wiki + web) &
3 \\

Qasper &
Long-document QA on NLP papers &
3 \\

NarrativeQA &
Generative narrative QA (stories) &
2 \\

GSM8K &
Grade-school math reasoning &
2 \\

Natural Questions &
Open-domain QA from search logs &
2 \\

FB15k-237 &
Knowledge-graph QA / link prediction &
2 \\

ALFWorld &
Text-based interactive environment &
2 \\

EntailmentBank &
Multi-step textual entailment proofs &
2 \\

MQuAKE-CF-3k / MQuAKE-T &
Counterfactual / transfer multi-hop QA &
2 \\

FEVER &
Wikipedia fact verification &
2 \\

CommonsenseQA &
Commonsense multiple-choice QA &
2 \\

OpenBookQA &
Open-book science QA &
2 \\
\bottomrule
\end{tabularx}
\caption{
Datasets that appear 2--3 times in the Topic~4 sample.
}
\label{tab:t4-datasets-tier2}
\end{table}

\section{Numbers of the our design axes}
\subsection{Overall Execution Plan}
\begin{table}[t]
\centering
\small
\begin{tabular}{lrr}
\toprule
\textbf{Scheduling pattern (Axis A)} & \textbf{\#papers} & \textbf{Share} \\
\midrule
Retrieve-then-Read retrieval only                   & 49 & 47.1\% \\
Interleaved only                             & 24 & 23.1\% \\
Interleaved + Plan-then-Execute              &  8 &  7.7\% \\
Retrieve-then-read + Plan-then-Execute              &  7 &  6.7\% \\
Interleaved + Test-time Search Scaling       &  7 &  6.7\% \\
Test-time Search Scaling only                &  5 &  4.8\% \\
Plan-then-Execute + Test-time Search Scaling &  1 &  1.0\% \\
Plan-then-Execute only                       &  1 &  1.0\% \\
Other / unspecified                          &  2 &  1.9\% \\
\midrule
\textbf{Total}                               & 104 & 100\% \\
\bottomrule
\end{tabular}
\caption{
Distribution of scheduling patterns (Axis~A) across all 104 coded
methods in our corpus (topic~T4: retrieval--reasoning coupling).
}
\label{tab:global-scheduling}
\end{table}
\FloatBarrier

\subsection{Index Structure}
\begin{table}[t]
\centering
\small
\begin{tabular}{lrr}
\toprule
\textbf{Index structure (Axis B)} & \textbf{\#papers} & \textbf{Share} \\
\midrule
Graph / KG / community      & 46 & 44.2\% \\
Flat / candidate list           & 29 & 27.9\% \\
Hierarchical / summary tree     &  3 &  2.9\% \\
Long-context                    &  3 &  2.9\% \\
Web / live retrieval                 &  6 &  5.8\% \\
Multi-modal / other                  &  5 &  4.8\% \\
Other / unspecified                  & 12 & 11.5\% \\
\midrule
\textbf{Total}                       & 104 & 100\% \\
\bottomrule
\end{tabular}
\caption{
Distribution of index (Axis~B) across all 104 methods.
Graph- and KG-style indices are the most common among retrieval--reasoning
coupling papers, followed by flat passage lists.
}
\label{tab:global-structure}
\end{table}

\FloatBarrier

\subsection{Next Step Execution Plan}
\begin{table}[t]
\centering
\small
\setlength{\tabcolsep}{2pt} 
\begin{tabularx}{\columnwidth}{Y r r}
% \begin{tabular}{lrr}
\toprule
\textbf{Control pattern (Axis C)} & \textbf{\#papers} & \textbf{Share}\\
\midrule
Policy-based  & 32 & 30.8\%\\
Rule-based & 31 & 29.8\%\\
Policy-based + Rule-based & 14 & 13.5\%\\
Rule-based + Verifier/PRM-based & 7 & 6.7\%\\
Policy-based  + Search-based & 5 & 4.8\%\\
Search-based + Verifier/PRM-based & 4 & 3.8\%\\
Search-based & 2 & 1.9\%\\
Verifier/PRM-based & 2 & 1.9\%\\
Unspecified & 2 & 1.9\%\\
Policy-based  + Verifier/PRM-based & 1 & 1.0\%\\
Policy-based  + Uncertainty-gated & 1 & 1.0\%\\
Rule-based + Search-based & 1 & 1.0\%\\
Rule-based + Uncertainty-gated & 1 & 1.0\%\\
Policy-based  + Rule-based + Verifier/PRM-based & 1 & 1.0\%\\
\midrule
\textbf{Total} & 104 & 100\%\\
\bottomrule
% \end{tabular}
\end{tabularx}
\caption{Distribution of next-step execution / control patterns across all 104 methods (Axis~C).}
\label{tab:axisC-control}
\end{table}
\FloatBarrier

\subsection{Dataset Coverage}
\begin{table}[t]
\centering
\small
\begin{tabular}{lrr}
\toprule
\textbf{Dataset} & \textbf{\#papers} & \textbf{Share in 104} \\
\midrule
HotpotQA                 & 35 & 33.7\% \\
2WikiMultiHopQA          & 14 & 13.5\% \\
MuSiQue                  & 12 & 11.5\% \\
ComplexWebQuestions      & 12 & 11.5\% \\
StrategyQA               &  6 &  5.8\% \\
IIRC                     &  1 &  1.0\% \\
FanOutQA                 &  1 &  1.0\% \\
\midrule
\textbf{Total methods}   & 104 & 100\% \\
\bottomrule
\end{tabular}
\caption{
Coverage of major multi-hop QA benchmarks in our coded corpus.
Counts are based on the \texttt{Datasets\_Summary} field in
Table~\ref{tab:global-scheduling}.
}
\label{tab:dataset-coverage}
\end{table}
\FloatBarrier
\subsection{Stop / Continue Criteria}

\begin{table}[t]
\centering
\small
\begin{tabular}{lrr}
\toprule
\textbf{Stop / continue pattern (Axis D)} & \textbf{\#papers} & \textbf{Share}\\
\midrule
Budget-based + Heuristic-done & 42 & 40.4\%\\
Budget-based & 40 & 38.5\%\\
Budget-based + Verifier-threshold & 10 & 9.6\%\\
Budget-based + Memory/Progress & 5 & 4.8\%\\
Heuristic-done & 5 & 4.8\%\\
Unspecified & 2 & 1.9\%\\
\midrule
\textbf{Total} & 104 & 100\%\\
\bottomrule
\end{tabular}
\caption{Distribution of stop / continue criteria across all 104 methods (Axis~D).}
\label{tab:axisD-stop}
\end{table}
\FloatBarrier

\section{Four Components Paper Numbers}
Table~\ref{tab:design-distribution} summarizes how the four components in our framework are instantiated across the 104 multi-hop QA / RAG papers in our Topic~4 sample. Along Axis~A (overall execution plan), more than half of the systems (56/104, 53.8\%) still follow a frozen-pool ``retrieve--then--read'' pattern, while 39/104 (37.5\%) explicitly
interleave retrieval and reasoning, 18/104 (17.3\%) implement
plan--then--execute strategies, and 13/104 (12.5\%) scale test-time search
(e.g., Tree-of-Thoughts or MCTS) on top of these patterns.

Axis~B (index structure) shows that graph- or KG-style indices are now
slightly more common (47/104, 45.2\%) than flat text indices
(33/104, 31.7\%), whereas hierarchical summary trees and long-context
retrieval units remain rare (3/104 each). A further 20 papers
(19.2\%) operate without a fixed static index, instead relying on web
search, multimodal retrieval, or purely parametric knowledge.

For Axis~C (next-step execute plan), over half of the systems employ
a learned global control policy (53/104, 51.0\%), and a similar number
still use rule-based action selection (56/104, 53.8\%) such as fixed
reasoning rhythms or hand-crafted heuristics; only 9/104 papers
(8.7\%) rely on fully search-based controllers (beam search, MCTS, or
Tree-of-Thoughts).

Finally, Axis~D (stop / continue criteria) is dominated by
resource-constrained stopping: 97/104 systems (93.3\%) impose explicit
hop, token, or latency budgets, and 47/104 (45.2\%) combine these with
simple ``done when the path is complete'' heuristics or beam-exhaustion
rules. Verifier- or PRM-based stopping remains less common but growing
(15/104, 14.4\%), and only 2 systems (1.9\%) explicitly use uncertainty-
or confidence-based stopping. These distributions support our claim that
most recent progress focuses on richer control and planning on top of
relatively conservative index structures and budgeted stopping schemes.

\begin{table}[t]
\centering
\small
\setlength{\tabcolsep}{2pt} 
\begin{tabularx}{\columnwidth}{Y r r}
\toprule
\textbf{Axis / choice} & \textbf{\#papers} & \textbf{Share of 104} \\
\midrule
\multicolumn{3}{l}{\textbf{(A) Overall execution plan}} \\
Retrieve--then--read & 56 & 53.8\% \\
Interleaved retrieval + reasoning  & 39 & 37.5\% \\
Plan--then--execute                & 18 & 17.3\% \\
Test-time search scaling (ToT/MCTS/beam) & 13 & 12.5\% \\
\midrule
\multicolumn{3}{l}{\textbf{(B) Index structure}} \\
Flat text / candidate list         & 33 & 31.7\% \\
Graph / KG / community index       & 47 & 45.2\% \\
Hierarchical / summary tree        & 3  & 2.9\% \\
Long-context units                 & 3  & 2.9\% \\
Other (web / multimodal / param.)  & 20 & 19.2\% \\
\midrule
\multicolumn{3}{l}{\textbf{(C) Next-step execute plan: control strategy}} \\
Policy-based control               & 53 & 51.0\% \\
Rule-based control                 & 56 & 53.8\% \\
Search-based control (beam / MCTS / ToT) & 9 & 8.7\% \\
\midrule
\multicolumn{3}{l}{\textbf{(D) Stop / continue criteria}} \\
Budget-based stopping (hop / token / latency) & 97 & 93.3\% \\
Heuristic / done-when-path-complete rules    & 47 & 45.2\% \\
Progress-based rules (memory / frontier)     & 5  & 4.8\% \\
Verifier- / PRM-based stopping               & 15 & 14.4\% \\
Uncertainty- / confidence-based stopping     & 2  & 1.9\% \\
\bottomrule
\end{tabularx}
\caption{
Distribution of retrieval--reasoning design choices across the 104
multi-hop QA / RAG papers in our Topic~4 sample, aligned with the four
components in Figure~\ref{fig:method-map}.
Counts are multi-label within each block: a single system may instantiate
multiple choices along an axis, so percentages do not sum to 100\%.
}
\label{tab:design-distribution}
\end{table}

\FloatBarrier     
\clearpage  
\onecolumn
\begin{longtable}{p{2.4cm}p{5.8cm}p{0.9cm}p{6.3cm}}
\toprule
\textbf{Dataset} & \textbf{System} & \textbf{Year} & \textbf{Reported performance on this dataset} \\
\midrule
\endfirsthead
\toprule
\textbf{Dataset} & \textbf{System} & \textbf{Year} & \textbf{Reported performance on this dataset} \\
\midrule
\endhead
\midrule
\multicolumn{4}{r}{\emph{Continued on next page}} \\
\midrule
\endfoot
\bottomrule
\endlastfoot

2WikiMultiHopQA &
Credible Plan-Driven RAG Method for Multi-hop Question Answering (PAR RAG) &
2025 &
2WikiMultiHopQA: EM 0.71, Acc 0.78. \\

2WikiMultiHopQA &
KG-o1: Enhancing Multi-hop Question Answering in LLMs with Knowledge Graphs &
2025 &
2WikiMultiHopQA: EM 62.40 / F1 74.45 / Prec. 72.56 / Rec. 79.45. \\

HotpotQA &
Resource-Friendly Dynamic Enhancement Chain for Multi-hop QA (DEC) &
2025 &
HotpotQA — CoverEM 47.19 / F1 50.96 / Acc\textsuperscript{\textdagger} 49.60;\dots\ HotpotQA — CoverEM 58.52 / F1 62.11 / Acc\textsuperscript{\textdagger} 60.32. \\

HotpotQA &
KG-o1: Enhancing Multi-hop Question Answering in LLMs with Knowledge Graphs &
2025 &
HotpotQA: EM 51.00 / F1 67.11 / Prec. 69.66 / Rec. 68.18. \\

MuSiQue &
End-to-End Beam Retrieval for Multi-Hop Question Answering &
2024 &
MuSiQue-Ans (test): Answer F1 = 69.2, Support F1 = 91.4 (comparable to human Sp = 93.9). \\

IIRC &
Interleaving Retrieval with Chain-of-Thought Reasoning for Knowledge-Intensive Multi-Step Questions (IRCoT) &
2023 &
Flan-T5-XXL: IRCoT vs OneR → +2.5 F1 on IIRC, with gains holding in OOD and at smaller model scales. \\

StrategyQA &
LLM-based Search Assistant with Holistically Guided MCTS for Intricate Information Seeking &
2025 &
StrategyQA: EM/F1 = 77.67 / 77.67 (GPT-4o-mini, best-performing setting). \\

FanOutQA &
LLM-based Search Assistant with Holistically Guided MCTS for Intricate Information Seeking &
2025 &
FanOutQA: Acc 58.38; R-1/2/L = 55.02 / 35.45 / 49.40 (outperforms other agent baselines). \\

\end{longtable}

\end{document}